\documentclass[11pt,logo,onecolumn,copyright]{nvidiatechreport}
\ifPDFTeX\else\microtypesetup{tracking=false}\fi
\usepackage[round]{natbib}
\usepackage{subfigure}
\usepackage{xspace}
\usepackage{placeins}
\usepackage{xcolor}
\usepackage{amsmath,amssymb}
\usepackage{booktabs}
\usepackage{multirow}

\newif\ifdraft
\drafttrue   

\ifdraft
  \newcommand{\prov}[1]{#1}
  \newcommand{\tbd}[1]{\textcolor{red}{[TBD: #1]}}
  \newcommand{\todo}[1]{\textcolor{blue}{[TODO: #1]}}
\else
  \newcommand{\prov}[1]{#1}
  \newcommand{\tbd}[1]{}
  \newcommand{\todo}[1]{}
\fi

\newcommand{\method}[0]{\textsc{AutoCompress}\xspace}

\newcommand{\avgeight}{Avg$_8$}

\newcommand{\algoName}[0]{\textsc{AutoCompress}\xspace}

\setlist[itemize]{topsep=2pt,itemsep=2pt}
\renewcommand{\today}{1 November 2026}
\title{AutoCompress: Automated Hardware-Guided Model Compression}
\author{Sheldon Liang, Ali Taghibakhshi\textsuperscript{*}, Saurav Muralidharan\textsuperscript{*},\\
{\Authfont Sharath Turuvekere Sreenivas, Ashwath Aithal, Pavlo Molchanov, Nima Tajbakhsh}\\
{\Affilfont \textsuperscript{*}Project Lead}}
\date{}
\hypersetup{
  pdftitle={AutoCompress: Automated Hardware-Guided Model Compression},
  pdfauthor={Sheldon Liang, Ali Taghibakhshi, Saurav Muralidharan, Sharath Turuvekere Sreenivas, Ashwath Aithal, Pavlo Molchanov, Nima Tajbakhsh},
  pdfsubject={NVIDIA Technical Report}
}

\begin{abstract}

LLM pruning must preserve quality while meeting hardware- and workload-specific constraints on latency, throughput, and memory.
Existing pruning workflows often target parameter counts, requiring practitioners to manually select model sizes that meet latency, throughput, and GPU memory limits. However, parameter count alone does not predict these metrics, which depend on the model architecture, workload, hardware, and serving engine.
We introduce \algoName, a hardware-guided compression framework that directly incorporates such constraints into differentiable architecture search.
\algoName jointly optimizes global and per-layer structure to produce heterogeneous architectures, guided by serving-engine cost models validated against end-to-end latency, throughput, and memory measurements. We evaluate \algoName on long-context local coding, batched datacenter serving, and latency-sensitive in-game interaction.
Compared to the state-of-the-art Minitron pruning framework, \algoName improves average accuracy across eight benchmarks by 6.57, 2.35, and 0.57 percentage points on RTX 5070, RTX 5080, and H100, respectively.
\algoName{} also compresses Nemotron 3.5 Super from 120B to 74B parameters for single-B200 BF16 serving at 128K context, retaining 95\% of its mean benchmark score.

\end{abstract}

\begin{document}
\maketitle
\begingroup
\renewcommand{\thefootnote}{}
\endgroup

\section{Introduction}
\label{sec:intro}

Large language models (LLMs) have demonstrated strong capabilities in reasoning, code generation, and long-context understanding. However, their substantial compute and memory requirements make them expensive to deploy, particularly on hardware with limited resources. 

In practice, the requirements for deploying an LLM vary considerably across applications and hardware. For instance, a local coding assistant must support long contexts within the limited memory of a consumer GPU, while a datacenter deployment must sustain high throughput when serving many requests concurrently. An in-game character presents a different challenge: the model must generate responses within a strict latency budget while sharing GPU resources with rendering. These scenarios place different demands on model compression, prioritizing memory capacity, batched throughput, or a combination of latency and memory limits. Effective compression therefore requires choosing an architecture that satisfies the specific deployment constraints while retaining as much of the original model's quality as possible.

\begin{figure}[!h]
\centering
\includegraphics[width=0.6\columnwidth]{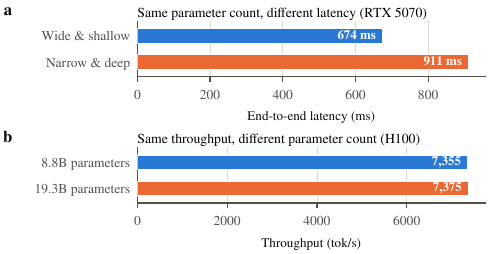}
\caption{\textbf{Parameter count is a poor proxy for serving cost.} (a)~Two model architectures with the same parameter count (\prov{3.61}B total, \prov{1.26}B active); the narrow, deep one has \prov{35}\% higher latency on an RTX~5070. (b)~Two model architectures that meet the same throughput on an H100; one has \prov{2.2}$\times$ the parameters of the other.}
\label{fig:params-latency}
\end{figure}

Existing pruning frameworks provide effective recipes for deriving smaller models, but adapting them to specific deployment requirements still requires substantial manual effort. For instance, Minitron~\cite{muralidharan2024compact,taghibakhshi2026efficient} performs architecture search within a user-defined search space and parameter budget, leaving practitioners to translate deployment constraints into suitable pruning targets. Additionally, parameter count alone does not capture the cost of executing a model on the target hardware and serving engine (see Figure~\ref{fig:params-latency}), and memory estimates based only on model weights and the KV cache omit additional allocations, such as execution workspaces and runtime buffers.

The search space is another important consideration: restricting layers to uniform dimensions, as in Minitron's search, excludes architectures that allocate capacity differently across layers. These considerations motivate an automated pruning framework that jointly searches model-wide and per-layer dimensions, supports explicit latency, throughput, and memory constraints, and uses cost estimates calibrated and validated within the target serving environment.

To address these challenges, we introduce \algoName, an automated, hardware-guided model pruning framework that uses differentiable architecture search to identify compressed models subject to latency or throughput targets, parameter budgets, and memory limits, either individually or in combination. The search jointly optimizes global architectural dimensions and per-layer structure, allowing different layers to retain different sizes and shapes. It uses cost models derived from measurements and allocation rules within the target serving engine.

\begin{figure*}[!h]
\centering
\includegraphics[width=\textwidth]{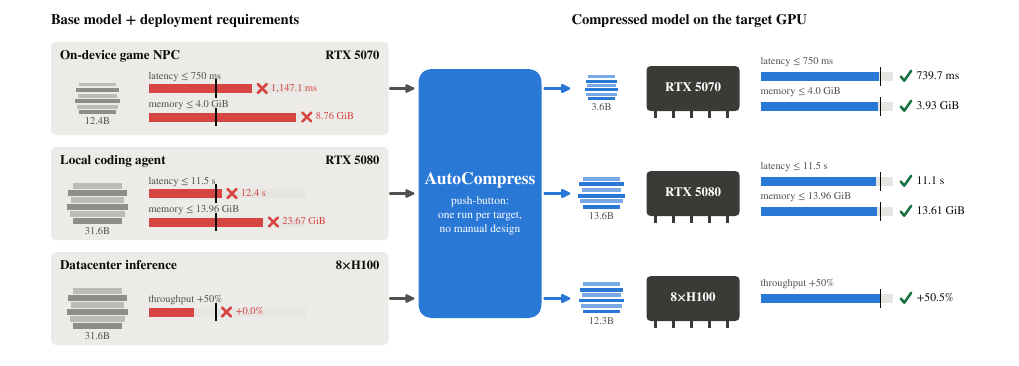}
\caption{
\textbf{\algoName{} overview.} Given a base model and deployment constraints, \algoName{} automatically selects, prunes, and distills an architecture for on-device gaming, local coding, or datacenter serving. The compressed models meet their latency, memory, or throughput targets, measured on real GPUs.
}
\label{fig:pipeline}
\end{figure*}

As illustrated in Figure~\ref{fig:pipeline}, \algoName{} takes a pretrained base model and a deployment target specifying the hardware, workload, and resource constraints, and produces a standalone compressed model. The process is fully automated, from cost modeling and architecture selection to pruning and quality recovery, eliminating manual architecture design and translation of deployment requirements into parameter budgets.
Across three distinct deployment scenarios, \algoName{} improves average accuracy over a deployment-adapted Minitron baseline by 6.57, 2.35, and 0.57 percentage points on RTX~5070, RTX~5080, and H100, respectively, while using approximately 2-3.3$\times$ fewer search tokens. 
We further demonstrate scalability by compressing Nemotron 3.5 Super from 120B to 74B parameters, enabling BF16 serving at 128K context and batch size 16 within 178\,GB on a single B200, with a 2.85-point decrease in mean benchmark score.

This paper makes the following key contributions:

\begin{itemize}
    \item \textbf{Hardware-guided compression.} We introduce \algoName, a differentiable search framework that jointly optimizes global and per-layer structure under latency, throughput, parameter, and memory constraints.

    \item \textbf{Serving-aware cost models.} We develop and validate cost models that capture execution time and memory allocations within the target serving engine.

    \item \textbf{Diverse deployment objectives.} We evaluate \algoName on long-context local coding, batched datacenter serving, and latency-sensitive in-game interaction against a deployment-adapted Minitron baseline.

    \item \textbf{Accuracy within budget.} \algoName improves average benchmark accuracy by 6.57, 2.35, and 0.57 percentage points on RTX 5070, RTX 5080, and H100, respectively, while meeting deployment targets.
    We further demonstrate scalability through a 120B-to-74B compression case study.
\end{itemize}

\section{Budget-Aware Architecture Search}
\label{sec:method}

\begin{figure*}[!h]
\centering
\includegraphics[width=\textwidth]{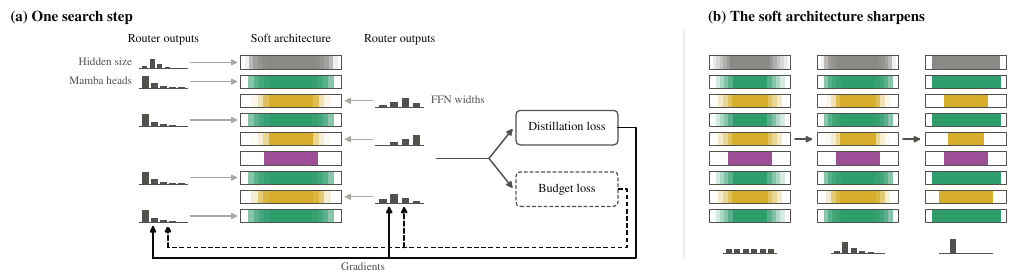}
\caption{\textbf{Joint search over global and per-layer structure.} (a) Router outputs define soft architecture masks, updated through gradients from distillation and budget losses. (b) As search progresses, the distributions sharpen toward discrete choices, yielding a heterogeneous architecture.}
\label{fig:router}
\end{figure*}


\algoName{} selects a compressed architecture for a specified deployment target by jointly training an architecture router and a weight-sharing supernet.
Given a pretrained parent and a search space, the router learns architectural choices that balance model quality against resource constraints. The selected architecture is then extracted from the parent and recovered through knowledge distillation.

\subsection{Deployment Target and Search Procedure}
\label{sec:contracts}

Given a pretrained model, which we refer to as the \emph{parent}, our goal is to derive a smaller model that satisfies a deployment target $\mathcal{D}$. $\mathcal{D}$ specifies the hardware, serving configuration, precision, and workload, including batch size, cached-prefix length, and input/output lengths. Resource constraints impose a memory limit, a latency limit or throughput requirement, or a combination of these.

Our procedure has four stages. First, we construct cost models for the target deployment (see \S\ref{sec:system} for more details). Second, we rank the parent's components by importance and reorder its weights to initialize a \emph{supernet}, which shares weights across candidate architectures. Third, we jointly train an architecture router, which selects among these candidates, and the supernet using distillation from the frozen parent and resource penalties. Finally, we extract the selected architecture from the reordered parent's weights and recover its quality through distillation, producing a standalone model.

\subsection{Search Space and Shared Weights}
\label{sec:space}
\label{sec:supernet}

Our search combines global dimensions shared across layers with per-layer dimensions that allow heterogeneous architectures. Candidate choices depend on the parent model, deployment target, and serving-engine support. We search over hidden size and Mamba head count, together with per-layer expert counts and routed and shared FFN widths. Mamba head count is shared across layers when varied globally. Supported per-layer choices also include zero width to remove a layer: for MoE layers, zero routed width results in a dense MLP when shared width is positive, while setting both widths to zero removes the layer. Thus, layer removal is represented within the same search space as width pruning. \S\ref{sec:search-spaces} specifies the candidate values and fixed dimensions for each deployment.

To share weights across candidates, we rank components by activation-based importance and permute their weights so that the highest-ranked components come first~\citep{muralidharan2024compact}. This ordering covers residual channels, Mamba heads within each head group, experts, and routed and shared FFN channels. Each candidate therefore uses a subset of the same supernet weights.




\subsection{Architecture Router}
\label{sec:budget}

The architecture router learns logits over the candidate values for each search axis, with separate logits for each layer on per-layer axes. As illustrated in Figure~\ref{fig:router}(a), these choices define a soft architecture optimized through distillation and budget losses. We sample differentiable architecture choices using Gumbel-softmax~\citep{jang2017gumbel,maddison2017concrete}:
\begin{equation}
  y_c = \frac{\exp((\kappa_t z_c + g_c)/\tau_t)}
  {\sum_j \exp((\kappa_t z_j + g_j)/\tau_t)}.
  \label{eq:gumbel}
\end{equation}
Here, $z_c$ is the logit for choice $c$, $g_c$ is independent standard Gumbel noise, and the sum spans all choices on the axis. The logit scale $\kappa_t$ controls the strength of learned preferences relative to noise, while the temperature $\tau_t$ controls how closely the relaxed choices approach one-hot selections. Increasing $\kappa_t$ and decreasing $\tau_t$ progressively sharpen the soft architecture toward discrete choices (Figure~\ref{fig:router}(b)); their schedules are specified in \S\ref{sec:search-training}.

\paragraph{Soft architecture masks.}
We translate each sampled choice into a mask over the importance-sorted components. A component's mask is the sum of $y_c$ over choices that retain it. For example, when choosing between widths 64 and 128, the first 64 channels receive weight one, while the remaining channels receive weight $y_{128}$. These masks allow gradients to update architecture choices while the shared weights adapt. Each batch is divided equally among independently sampled configurations. Expert-count masks additionally account for top-$k$ expert selection, as described in \S\ref{sec:expert-masking}.


\paragraph{Joint objective.}
The frozen parent serves as the distillation teacher, while the router and supernet are trained jointly. We combine logit distillation~\citep{hinton2015distilling} with a resource penalty:
\begin{equation}
  \mathcal{L} = \mathcal{L}_{\mathrm{KD}} + \alpha \sum_r
  \left[\max(0,\rho_r-1)+\max(0,1-\delta-\rho_r)\right].
  \label{eq:loss}
\end{equation}
Here, $\mathcal{L}_{\mathrm{KD}}$ is the forward KL divergence from the parent's token distribution to the sampled supernet's distribution. For each constrained resource $r$, $\rho_r$ is the cost predicted from the soft architecture choices divided by its budget. For example, a latency cap $T_{\max}$ gives $\rho_T=\hat{T}/T_{\max}$. At a fixed workload, a throughput requirement is expressed as an upper bound on serving time.

The coefficient $\alpha$ balances resource constraints against distillation, while $\delta$ controls the permitted unused budget. We use $\delta=0.02$, yielding zero penalty between 98\% and 100\% of each cap. The penalty discourages budget violations above this band and encourages greater capacity below it.

\subsection{Architecture Extraction and Recovery}
\label{sec:extract}

Once architecture search converges, we select the highest-logit choice from each router gate and evaluate the resulting discrete architecture using our cost models. We initialize the selected model with the corresponding weights from the importance-sorted parent and recover its quality through logit distillation (\S\ref{sec:recovery-training}). Recovery thus starts from the retained parent weights, not the trained supernet weights.

\section{Modeling Deployment Cost}
\label{sec:system}

\begin{figure*}[tb]
\centering
\includegraphics[width=\textwidth]{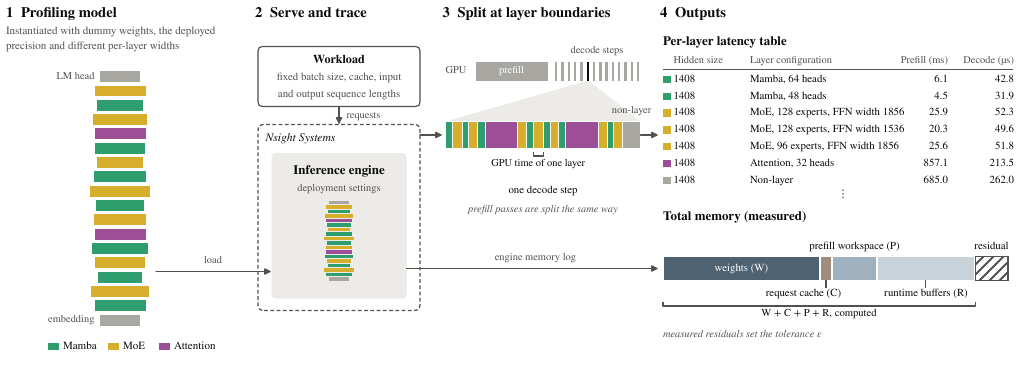}
\caption{\textbf{Measuring per-layer latency and memory} (1)~A model with multiple different layer configurations is instantiated. (2)~Inference engine serves the model under the deployment setting while Nsight Systems traces runtime. (3)~The trace is split at layer boundaries, for prefill and for every decode step. (4)~Each layer's runtime becomes one row of the latency table; One profiling run obtains many layer measurements along with total memory usage, whose residual over $W + C + P + R$ (Eq.~\ref{eq:memory}) sets the tolerance $\epsilon$.}
\label{fig:measurement}
\end{figure*}


\algoName{} uses cost models to estimate each search candidate's latency, throughput, and memory usage under the target serving configuration. We describe these models below and evaluate their accuracy and profiling cost in \S\ref{sec:validation} and \S\ref{sec:overhead}, respectively.

\subsection{Latency and throughput}
\label{sec:tables}

Profiling every complete architecture is impractical due to the large combinatorial space produced by joint global and per-layer choices.
We instead measure individual layer configurations within the target serving engine and combine their runtimes to estimate model latency. A deployment scenario $\mathcal{D}$ specifies the hardware, serving configuration, precision, and workload, including concurrency, batch size, and input and output lengths. For architecture $a$, we estimate latency as
\begin{equation}
  \hat{T}(a \mid \mathcal{D})
  = T_0(d \mid \mathcal{D})
  + \sum_{\ell=1}^{L} t_\ell(a_\ell; d \mid \mathcal{D}).
  \label{eq:latency}
\end{equation}
Here, $d$ is the global hidden size, $L$ is the parent's layer count, and $a_\ell$ specifies the configuration at layer $\ell$. The measured runtime $t_\ell$ is zero for a removed layer, while $T_0$ accounts for embedding, the LM head, sampling, and serving-engine overhead.

\paragraph{Measurement procedure.}
Each profiling run serves a short model containing approximately ten different layer configurations initialized with dummy weights. We match the target deployment's engine, precision, batch size, sequence lengths, and cache settings, and use Nsight Systems traces to measure GPU time at layer boundaries separately for prefill and decode. A single run therefore populates multiple entries in the timing table (Figure~\ref{fig:measurement}). We estimate each candidate's latency by summing its corresponding entries and adding $T_0$.

Profiling within the serving engine captures deployment-specific kernels and execution behavior. However, architectural changes can affect kernel selection, cache behavior, and overhead, limiting the accuracy of an additive estimate. We validate these estimates on complete architectures excluded from timing-table construction (\S\ref{sec:validation}).

\subsection{Memory}
\label{sec:memory}

For architecture $a$ under deployment scenario $\mathcal{D}$, we estimate serving memory as
\begin{equation}
  \hat{M}(a \mid \mathcal{D}) = W + C + P + R + \epsilon_{\mathcal{D}}.
  \label{eq:memory}
\end{equation}
Here, $W$ is weight memory computed from tensor shapes and deployed data types. The request cache $C$ includes attention KV and Mamba convolution and SSM caches, sized for the target sequence length and batch size. The prefill workspace $P$ depends on the query-head configuration and prefill chunk size, while runtime buffers $R$ include MoE matrix-multiplication and Mamba projection workspaces sized by the relevant widest layers.

The allowance $\epsilon_{\mathcal{D}}$ accounts for CUDA context, graph allocations, and other unmodeled memory. We set it to the largest residual between measured memory and $W+C+P+R$ across calibration server starts under $\mathcal{D}$. These measurements can be collected during latency profiling without additional runs (Figure~\ref{fig:measurement}).




\section{Experimental Setup}
\label{sec:setup}



We evaluate \algoName{} across three deployment scenarios with distinct latency, throughput, and memory requirements. Below, we describe the models and workloads, search spaces, baseline, and training and evaluation procedures.

\subsection{Models and Deployment Settings}
\label{sec:deployment-setup}

We use two parent models with hybrid Mamba, attention, and MoE architectures: Nemotron Nano 3 Elastic 12B~\cite{taghibakhshi2026efficient} and Nemotron 3.5 Lightning~\cite{NVIDIA_Nemotron_3_5_Super_Lightning}. Table~1 summarizes the workloads and resource limits for each scenario.

\paragraph{On-device game NPC.}
An interactive character must respond quickly while sharing GPU memory with rendering. We compress Nemotron Nano 3 Elastic 12B for an RTX 5070, the current most popular GPU among Steam users~\cite{steam2026survey}. We impose a 4\,GiB memory budget, leaving 8\,GiB of memory, equivalent to an entry-level gaming GPU, for game rendering. Latency is targeted at 750ms end-to-end to match modern voice agent standards~\cite{nvidia2026voiceagent}. The workload, 2048 input and 256 output tokens, is taken as an upperbound of sample lengths in the NPC dialogue V2 dataset ~\cite{npc_dialogue_v2_2024}.

\paragraph{Local coding agent.}
A local coding agent must process long contexts within consumer-GPU memory limits. We compress Nemotron 3.5 Lightning for an RTX~5080 using a long cached prefix followed by additional input and single-request generation. Workload and latency targets come from TraceLab~\cite{zhu2026tracelab}, a trace dataset of production coding agents. We round up recorded Codex p90 sequence lengths to set our workload, 262{,}144 cache, 16{,}384 input, and 1{,}024 output tokens. The latency target is 11.5\,s, the mean time an LLM step takes in the dataset traces. The memory target is set as the limit of vLLM ~\cite{kwon2023vllm} running on an RTX~5080 GPU with default memory utilization, 13.96\,GiB.

\paragraph{Datacenter inference.}
Batched serving prioritizes aggregate throughput. We use the same Lightning parent on an 8$\times$H100 node with a decode-heavy workload of 8{,}192 input and 65{,}536 output tokens per request. After sweeping parallelism layouts and batch size, TP1 $\times$ DP8 with batch size 128 per GPU maximizes the node's aggregate throughput. We then set the target of 7{,}330 output tokens per second per GPU, a 50\% increase of throughput over the parent model.


All models are served using vLLM 0.27.1~\citep{kwon2023vllm}. Profiling and cost estimation use the same precision, workload, and cache settings as deployment (see Appendix~\ref{sec:apx-serving} for details).



\begin{table*}[t]
\centering
\small
\caption{Deployment settings.}
\label{tab:contracts}
\begin{tabular}{@{}p{0.14\textwidth}p{0.26\textwidth}p{0.26\textwidth}p{0.26\textwidth}@{}}
\toprule
 & \textbf{On-device game NPC} & \textbf{Local coding agent} & \textbf{Datacenter inference} \\
\midrule
Hardware & 1$\times$ RTX 5070 (12\,GB) & 1$\times$ RTX 5080 (16\,GB) & 8$\times$ H100, TP1 $\times$ DP8 \\
Parent & Nemotron Nano 3 Elastic 12B (\prov{12.4}B / \prov{2.1}B active) & Nemotron 3.5 Lightning (\prov{31.6}B / \prov{3.6B active}) & Nemotron 3.5 Lightning (\prov{31.6}B / \prov{3.6B active}) \\
Workload & batch size 1; \prov{2{,}048} input + \prov{256} output tokens & batch size 1; \prov{262{,}144} cached + \prov{16{,}384} input + \prov{1{,}024} output tokens & batch size 128 per GPU; \prov{8{,}192} input + \prov{65{,}536} output tokens \\
Latency / throughput target & latency $\le$ \prov{750}\,ms & latency $\le$ \prov{11.5}\,s & throughput $\ge$\prov{7{,}330} tok/s per GPU (+50\% over the parent) \\
Memory target & $\le$\prov{4.0}\,GiB & $\le$\prov{13.96}\,GiB & -- \\
Parent cost & \prov{1{,}147}\,ms, \prov{8.76}\,GiB & \prov{12.44}\,s, \prov{23.67}\,GiB & \prov{4{,}922} tok/s per GPU \\
\bottomrule
\end{tabular}
\end{table*}


\subsection{Candidate Architectures}
\label{sec:search-spaces}


Table~\ref{tab:search-space} lists the candidate values for each deployment setting. With 23 Mamba and 23 MoE layers in each parent, these choices yield approximately $2.37 \times 10^{42}$ distinct architectures for Lightning 3.5 and $1.20 \times 10^{52}$ for Nano 3 12B before applying deployment constraints. 

\begin{table*}[t]
\centering
\small
\setlength{\tabcolsep}{5pt}
\caption{Candidate values for learned architecture search. Hidden size is global; expert counts and FFN widths vary by layer.}
\label{tab:search-space}
\begin{tabular}{@{}lll@{}}
\toprule
\textbf{Dimension} & \textbf{Lightning 3.5: RTX 5080 / H100} & \textbf{Nano 3 12B: RTX 5070} \\
\midrule
Hidden size $d$ & 2688, 2432, 2176, 1920, 1664, 1408 & 1920, 1792, 1664, 1536, 1408, 1280 \\
Mamba  $H$ & 64, 56, 48, 40, 32 (global) & 64 or remove (per layer) \\
Routed experts $E_\ell$ & 128, 112, 96, 80, 64 & 128, 96, 64, 32 \\
Routed FFN width $F_\ell$ & 1856, 1536, 1280, 1024 & 960, 704, 448, 192, 0 \\
Shared FFN width $S_\ell$ & 3712, 2816, 1920 & 3712, 2688, 1856, 1024, 0 \\
Depth & Fixed & Mamba and MoE layers may be removed \\
\bottomrule
\end{tabular}

\end{table*}

\subsection{Minitron++ Baseline}
\label{sec:baseline}

Existing pruning methods do not directly support our joint search space and deployment constraints. We therefore adapt the Minitron pruning and screening procedure~\citep{muralidharan2024compact} to use our serving-cost models, with the same importance-sorted parent and deployment budgets as \algoName{}. We call this new baseline Minitron++. We use uniform per-layer dimensions and retain all layers. We enumerate combinations from Table~\ref{tab:search-space}, except on RTX~5070, where Mamba head counts range from 64 to eight in steps of eight and zero FFN widths are excluded.

We retain candidates within 5\% below every resource cap that cannot be enlarged along any axis without shrinking another or violating a constraint. This yields \prov{6}, \prov{11}, and \prov{95} candidates for RTX~5070, RTX~5080, and H100, respectively. We use a wider band than \method{}'s 2\% because the latter admits only \prov{one} maximal RTX~5080 candidate.

Each candidate receives 100 distillation iterations (approximately \prov{1.7}B tokens), using the target's batch size, data, and peak learning rate. We select the candidate with the lowest mean loss over iterations 76--100. For H100, we first rank all 95 candidates by loss before training and screen only the top five. The selected architecture receives the same recovery training as \algoName{}'s student.





\subsection{Search Training Details}
\label{sec:search-training}

\paragraph{Router and optimization.}
Each router gate is a two-layer network with hidden width 256. Its output layer is zero-initialized, while the first layer is randomly initialized. The router uses a learning rate of $2 \times 10^{-3}$ without warmup. The supernet uses peak and minimum learning rates of $4 \times 10^{-5}$ and $5 \times 10^{-6}$, with a 60-step warmup. Each batch contains 2048 sequences of 8192 tokens, divided equally among 32 sampled configurations.

\paragraph{Sampling schedule and termination.}
Over 250 steps, the logit scale in Eq.~\ref{eq:gumbel} increases linearly from 1 to 10, while the temperature decreases geometrically from 1.0 to 0.05. Approximately 0.4\% of samples use hard one-hot choices. We extract an architecture once every gate assigns at least 0.95 probability to its leading choice. The reported architectures were extracted at iterations 170, 249, and 179 for RTX~5080, RTX~5070, and H100, respectively.

\paragraph{Budget-weight selection.}
The resource penalty is zero between 98\% and 100\% of each cap. We selected $\alpha=0.3$ through an H100 sweep and reused it for RTX~5080. RTX~5070 uses a higher budget weight of $\alpha=1$ due to its tighter budget constraints.

\paragraph{Parallelism.}
RTX~5080- and H100-targeted searches use eight-way tensor parallelism and eight-way expert parallelism across 32 H100 nodes. The RTX~5070-targeted search uses four-way tensor parallelism and eight-way expert parallelism across 8 H100 nodes. Search costs are reported in \S\ref{sec:overhead}.

\subsection{Distillation Details}
\label{sec:recovery-training}


After architecture selection, we use knowledge distillation to recover quality lost during pruning. Each student is sliced from the importance-sorted parent and trained using only the parent's output logits as supervision. For each deployment target, \algoName{} and baseline students receive the same data in the same order, batch size, and learning-rate schedule. Batches contain 2048 sequences of 8192 tokens, with peak and minimum learning rates of $4 \times 10^{-5}$ and $5 \times 10^{-6}$, respectively.

\subsection{Benchmarks}
\label{sec:eval-protocol}

We evaluate reasoning, coding, knowledge, and instruction following across eight benchmarks: AIME25~\cite{aime2025}, LiveCodeBench v5~\cite{jain2025livecodebench}, MMLU-Pro~\cite{wang2024mmlupro}, MATH-500~\cite{hendrycks2021math,lightman2024verify}, GSM8K~\cite{cobbe2021gsm8k}, IFEval~\cite{zhou2023ifeval}, IFBench~\cite{pyatkin2025ifbench}, and GPQA-Diamond~\cite{rein2024gpqa}. All evaluations use reasoning mode with temperature 1.0 and top-$p$ 1.0. We sample 16 responses per problem for AIME25, eight for GPQA-Diamond, one for MMLU-Pro, and four for the remaining benchmarks. We report pass@1 averaged over samples, with prompt-level strict accuracy for IFEval and IFBench.


\section{Results}
\label{sec:results}

\begin{figure*}[tb]
\centering
\includegraphics[width=\textwidth]{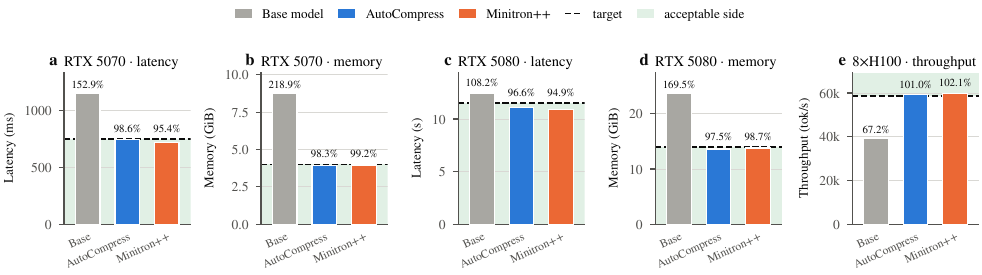}
\caption{\textbf{Budget adherence.} All selected architectures satisfy their deployment budgets when measured on real hardware.}
\label{fig:budget}
\end{figure*}

We compare \method{} with the Minitron++ homogeneous baseline under the same deployment budgets and matched recovery training.

\subsection{Accuracy Under Deployment Budgets}
\label{sec:quality}
\label{sec:fit}

As seen from Table~\ref{tab:quality}, \method{} improves \avgeight{} by \prov{6.57}, \prov{2.35}, and \prov{0.57} percentage points for RTX~5070, RTX~5080, and H100, respectively. Final distillation loss is lower by \prov{19.3}\%, \prov{4.3}\%, and \prov{0.6}\%, respectively.

\begin{table*}[tb]
\centering
\small
\caption{\textbf{Accuracy and architecture-selection tokens under matched deployment budgets.} Search tokens count architecture selection; KD tokens count recovery training. KD loss is averaged over the final 25 iterations on matched batches. Benchmark scores are final-checkpoint pass@1 (\%), with prompt-level strict accuracy for IFEval and IFBench. Evaluation follows \S\ref{sec:eval-protocol}; \avgeight{} averages the eight benchmarks.}
\label{tab:quality}
\resizebox{\textwidth}{!}{%
\begin{tabular}{@{}ll rr c rrrrrrrr r@{}}
\toprule
Contract & Method & Search tokens $\downarrow$ & KD tokens & KD loss $\downarrow$ & AIME25 & LiveCodeBench & MMLU-Pro & MATH-500 & GSM8K & IFEval & IFBench & GPQA & \avgeight{} \\
\midrule
RTX 5070 & \method{} & \textbf{\prov{4.18}B} & \prov{30}B & \textbf{\prov{0.1909}} & \textbf{\prov{0.42}} & \textbf{\prov{1.11}} & \textbf{\prov{26.46}} & \textbf{\prov{36.40}} & \textbf{\prov{47.54}} & \textbf{\prov{55.13}} & \textbf{\prov{24.49}} & \textbf{\prov{21.78}} & \textbf{\prov{26.67}} \\
 & Minitron++ & \prov{8.49}B & \prov{30}B & \prov{0.2366} & \prov{0.00} & \prov{0.48} & \prov{19.76} & \prov{29.90} & \prov{34.29} & \prov{43.76} & \prov{17.43} & \prov{15.21} & \prov{20.10} \\
\midrule
RTX 5080 & \method{} & \textbf{\prov{2.85}B} & \prov{10}B & \textbf{\prov{0.1637}} & \textbf{\prov{52.71}} & \textbf{\prov{49.44}} & \prov{67.45} & \textbf{\prov{93.60}} & \prov{92.87} & \textbf{\prov{83.64}} & \textbf{\prov{55.61}} & \prov{55.87} & \textbf{\prov{68.90}} \\
 & Minitron++ & \prov{8.57}B & \prov{10}B & \prov{0.1710} & \prov{48.75} & \prov{43.81} & \textbf{\prov{67.89}} & \prov{92.70} & \textbf{\prov{92.93}} & \prov{79.57} & \prov{49.83} & \textbf{\prov{56.94}} & \prov{66.55} \\
\midrule
8$\times$H100 & \method{} & \textbf{\prov{3.00}B} & \prov{10}B & \textbf{\prov{0.1809}} & \textbf{\prov{55.42}} & \textbf{\prov{48.02}} & \textbf{\prov{64.80}} & \prov{92.80} & \prov{92.49} & \textbf{\prov{82.53}} & \textbf{\prov{53.74}} & \prov{53.09} & \textbf{\prov{67.86}} \\
 & Minitron++ & \prov{9.98}B & \prov{10}B & \prov{0.1820} & \prov{53.54} & \prov{47.78} & \prov{64.43} & \textbf{\prov{92.85}} & \textbf{\prov{92.72}} & \prov{81.38} & \prov{52.04} & \textbf{\prov{53.60}} & \prov{67.29} \\
\bottomrule
\end{tabular}}

\end{table*}

All selected architectures meet their predicted deployment budgets (Figure~\ref{fig:budget}). The RTX~5070 and RTX~5080 \method{} students also meet their latency and memory limits on hardware, measuring \prov{739.7}\,ms / \prov{3.93}\,GiB and \prov{11.1}\,s / \prov{13.61}\,GiB, respectively.

\subsection{Architecture Selection Cost}
\label{sec:search-cost}

The Minitron++ baseline uses approximately \prov{2.0}$\times$, \prov{3.0}$\times$, and \prov{3.3}$\times$ as many architecture-selection tokens as \method{} for RTX~5070, RTX~5080, and H100, respectively (Table~\ref{tab:quality}). Recovery training uses the same token budget for both methods within each target.

\subsection{Case Study at Scale}
\label{sec:scale-case-study}

\begin{figure*}[!t]
\centering
\includegraphics[width=0.96\textwidth]{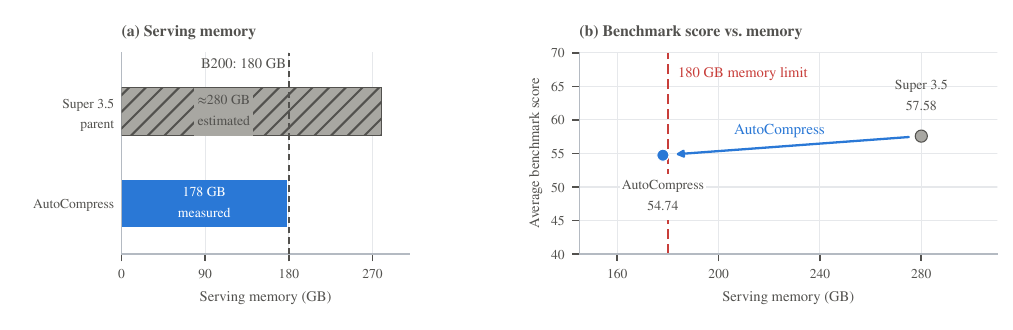}
\caption{\textbf{Compression at scale.}
(a) The Super 3.5 parent's estimated BF16 serving footprint of approximately 280\,GB exceeds the B200's 180\,GB capacity; the \method{} model supports a maximum batch size of 16 at 131{,}072-token context, with measured serving memory of 178\,GB.
(b) Mean benchmark score versus serving memory; the arrow connects the parent to the \method{} model, and the red dashed line marks the 180\,GB memory limit. The mean includes 9 benchmarks and the breakdown is available in Appendix~\ref{app:scale-details}.}
\label{fig:scale-memory-architecture}
\end{figure*}

To test \algoName{} at a larger scale, we set the deployment objective to fit Nemotron 3.5 Super~\citep{NVIDIA_Nemotron_3_5_Super_Lightning} on a single B200 GPU in BF16 at a context length of 131{,}072 tokens. The parent's estimated serving footprint is approximately \prov{280}\,GB, exceeding the device's 180\,GB capacity. At this context length, the \method{} model supports a maximum batch size of \prov{16}, using \prov{178}\,GB of GPU memory (Figure~\ref{fig:scale-memory-architecture}a). The resulting architecture has approximately \prov{74}B parameters, down from \prov{120}B in the parent, a \prov{38.3}\% reduction.

Architecture selection uses deployment memory footprint constraint, and the router selects layer-specific MoE routed expert counts and intermediate sizes while retaining the parent's backbone layers. Appendix~\ref{app:scale-details} gives the final architecture and individual benchmark scores.

Architecture search converges within \prov{4}B tokens, in similar range to previous smaller scale use in Table~\ref{tab:quality}. Recovery uses \prov{68}B tokens at 8192 context and \prov{48}B at 131072, for \prov{120}B tokens in total. Search and recovery used \prov{32} GB200 nodes (\prov{128} GPUs) for approximately \prov{64} hours. Given the training cost, we did not run the Minitron++ baseline for this case study.

After recovery, the unweighted mean of the nine reported benchmark scores is \prov{54.74}, compared with \prov{57.58} for the parent, a decrease of \prov{2.85} points (Figure~\ref{fig:scale-memory-architecture}b). \method{} also exceeds the \prov{45.44} mean of Nemotron 3 Super, a 120B-parameter reference model, by \prov{9.30} points.

\section{Ablations and Analysis}
\label{sec:analysis}

We examine how router initialization and expert masking affect architecture selection and model quality. We then analyze the search dynamics and learned architectures to understand how distillation and resource constraints shape global and per-layer choices.

\subsection{Router Initialization}
\label{sec:router-ablations}
\label{sec:initialization}

Random router-output initialization creates an initial bias over architecture choices. Learning-rate warmup limits the early updates that could correct this bias before the sampling distribution sharpens. We therefore initialize the router's output layer to zero and remove router warmup, while retaining warmup for the model weights.
In the RTX~5080 comparison, this configuration selects an architecture with \prov{10.9}\% lower distillation loss than random router-output initialization with warmup, as shown in Figure~\ref{fig:router-prior}.

\IfFileExists{figures/f3c_startup_kd.pdf}{%
\begin{figure}[tb]
\centering
\includegraphics[width=0.6\columnwidth]{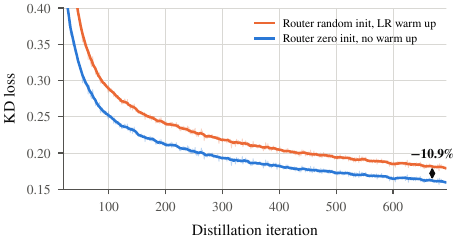}
\caption{\textbf{Router initialization and warmup.} Distillation loss of architectures selected by two RTX~5080 searches. Zero router-output initialization without router warmup yields \prov{10.9}\% lower loss than random initialization with warmup.}
\label{fig:router-prior}
\end{figure}
}{}

\subsection{MoE Routed Expert Masking}
\label{sec:expert-masking}

\begin{figure}[t]
\centering
\includegraphics[width=0.6\columnwidth]{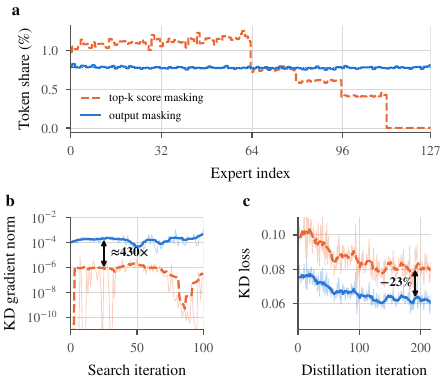}
\caption{\textbf{MoE routed expert masking.} Comparison of top-k score masking and expert output masking in expert token share, router KD gradient and architecture quality.}
\label{fig:expert-masking}
\end{figure}

A direct extension of channel masking to expert masking is applying soft masks to expert routing scores before top-$k$ selection. We have found that this reduces traffic to experts retained only by larger architecture choices. These experts then contribute little gradient to the architecture router. As an alternative, output masking applies the soft mask after expert selection, preserving access to these experts during search.
We compare the 2 expert masking methods in Figure~\ref{fig:expert-masking}. Output masking ensures evenly routed experts and provides stable distillation gradients. The resulting architecture gets $23\%$ lower distillation loss compared to using top-k score masking.

\subsection{Router Decisions and Learned Architectures}
\label{sec:gradsplit}
\label{sec:dynamics}

We log architecture-choice probabilities and gradients from the distillation and budget terms of Eq.~\ref{eq:loss}. Figure~\ref{fig:kd-budget-split} shows the RTX~5080 search. The shared-expert intermediate size settles first and remains at full size, followed by Mamba heads at iteration \prov{41} and hidden size at iteration \prov{50}. MoE routed expert counts and intermediate sizes settle last. Distillation-gradient norms are larger for the shared and global choices, while budget-gradient norms are larger for the remaining MoE choices. Preserving the shared expert is consistent with its relatively low serving cost: at hidden size 2688, nearly halving its intermediate size saves only \prov{7.5}\,ms and \prov{5}\,MiB per layer, compared with \prov{33.7}\,ms and \prov{307}\,MiB for a similar reduction in MoE intermediate size.

\IfFileExists{figures/f12_router_rtx5080.pdf}{%
\begin{figure}[t]
\centering
\includegraphics[width=0.6\columnwidth]{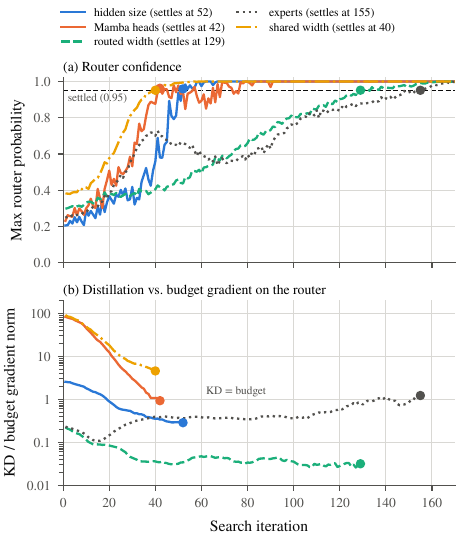}
\caption{\textbf{RTX~5080 router diagnostics.} Top: maximum choice probability on each architecture axis. Bottom: ratio of distillation-gradient norm to budget-gradient norm; a value above one indicates a larger distillation-gradient norm.}
\label{fig:kd-budget-split}
\label{fig:convergence}
\end{figure}
}{%
\begin{figure}[t]
\centering
\includegraphics[width=0.6\columnwidth]{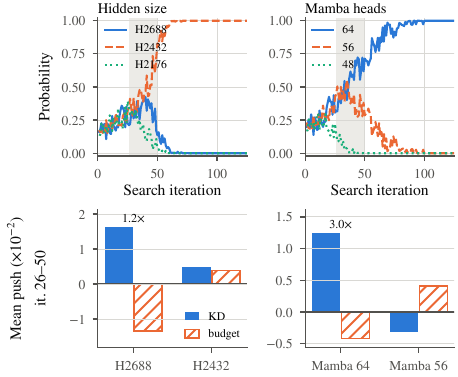}
\caption{\textbf{RTX~5080 global router choices.} Top: probabilities of hidden-size and Mamba-head choices. Bottom: mean signed contributions from distillation and budget gradients over iterations 26--50.}
\label{fig:kd-budget-split}
\label{fig:convergence}
\end{figure}
}

On H100, \method{} and Minitron++ select the same hidden size and Mamba head count. Minitron++ uses uniform MoE routed expert counts and intermediate sizes, while \method{} varies both across layers. At the same throughput target, \method{} achieves \prov{0.6}\% lower distillation loss and \prov{0.57} percentage points higher \avgeight{} (Table~\ref{tab:quality}). These gains come with a \prov{3.3}$\times$ reduction in architecture-selection tokens relative to Minitron++. Complete architectures are given in Appendix~\ref{app:layers}.
\section{Related Work}
\label{sec:related}

\paragraph{Structured LLM pruning.}

Structured pruning reduces model size by removing layers, attention heads, or hidden and intermediate dimensions~\citep{xia2022structured,ma2023llm,xia2024sheared,ashkboos2024slicegpt,men2025shortgpt}, with recent work extending pruning and recovery to large MoE models~\citep{tang2026slimqwen}. Minitron~\citep{muralidharan2024compact} combines importance-based pruning with knowledge distillation; we follow this recipe to initialize our supernet and adapt its architecture-selection procedure as our evaluation baseline. These methods primarily optimize retained structure and quality recovery at a prescribed model size or compression ratio.

Hardware-aware approaches incorporate execution cost into architecture selection. ZipLM~\citep{kurtic2023ziplm} iteratively removes components according to their loss--runtime trade-off, while SPRINT~\citep{park2025accurate} selects sublayers using latency reduction and post-pruning tunability. Related approaches include latency-constrained knapsack optimization~\citep{shen2022structural}, hardware-in-the-loop architecture search for mobile deployment~\citep{huang2026mobilellm}, and joint optimization of latency and model size~\citep{ali2026multi}.
Puzzle~\citep{bercovich2024puzzle} is particularly relevant: it 
evaluates candidate block replacements using blockwise distillation, selects a heterogeneous architecture through mixed-integer programming under throughput and memory constraints, and recovers quality through global distillation. Its formulation selects per-layer replacements at a fixed hidden dimension and does not support hidden-dimension pruning. The hybrid MoE extension introduces iterative compression, heterogeneous expert widths and routing budgets, and Mamba state pruning~\citep{bercovich2026nemotron}; however, the public implementation does not include Mamba head or expert-width pruning. In contrast, \algoName{} jointly optimizes hidden size, Mamba head count, and per-layer MoE structure through differentiable search, guided by serving-engine cost models calibrated to the target hardware, workload, and precision.

\paragraph{Cost models and serving.}
Prior work highlights the need to validate latency predictors for both accuracy and architecture ranking~\citep{akhauri2024latency,zhang2021nnmeter}. MoE serving costs also depend on execution phase and parallelism, which FLOPs alone do not capture~\citep{patel2026demystifying}. Complementary approaches optimize runtime memory use~\citep{ukarande2026efficient} or serving configurations for a fixed model~\citep{cho2026optimizing}. We instead guide architecture search with cost models derived from serving-engine measurements and allocation rules~\citep{kwon2023vllm}, using the target workload and precision, and validate their predictions on held-out models running on the target hardware (\S\ref{sec:validation}).


\section{Conclusion}
\label{sec:conclusion}

We have presented \algoName{}, a hardware-guided compression framework that jointly optimizes global and per-layer structure under latency, throughput, parameter, and memory constraints. Its cost models capture execution time and memory allocations within the target serving engine, accounting for the hardware and workload used in deployment. A budget-aware architecture router combines these cost estimates with distillation signals to learn how much capacity to retain across layers. The selected architecture is extracted from the parent and recovered through distillation, producing a standalone model without manual architecture design.

Across long-context local coding, batched datacenter serving, and latency-sensitive in-game interaction, \algoName{} improves accuracy over the deployment-adapted Minitron++ baseline. With matched recovery training, \avgeight{} increases by \prov{6.57}, \prov{2.35}, and \prov{0.57} percentage points on RTX~5070, RTX~5080, and H100, respectively, while architecture-selection tokens decrease by factors of approximately \prov{2.0}, \prov{3.0}, and \prov{3.3}.
All selected architectures meet their predicted deployment budgets, with serving measurements confirming latency and memory adherence for both RTX \algoName{} students.
The 120B-to-74B Nemotron 3.5 Super case study extends these results to larger models, enabling single-GPU BF16 serving at 128K context while retaining approximately 95\% of the parent's mean benchmark score.

\bibliography{refs}
\bibliographystyle{mlsys2025}

\clearpage
\appendix

\section{Per-Layer Architectures}
\label{app:layers}
\begin{figure*}[t!]
\centering
\small
\caption{\textbf{Architectures.} Both parent models have 52 total layers, 23 Mamba layers, 23 MoE layers and 6 attention layers. Attention layers are all kept full.}
\includegraphics[width=\textwidth]{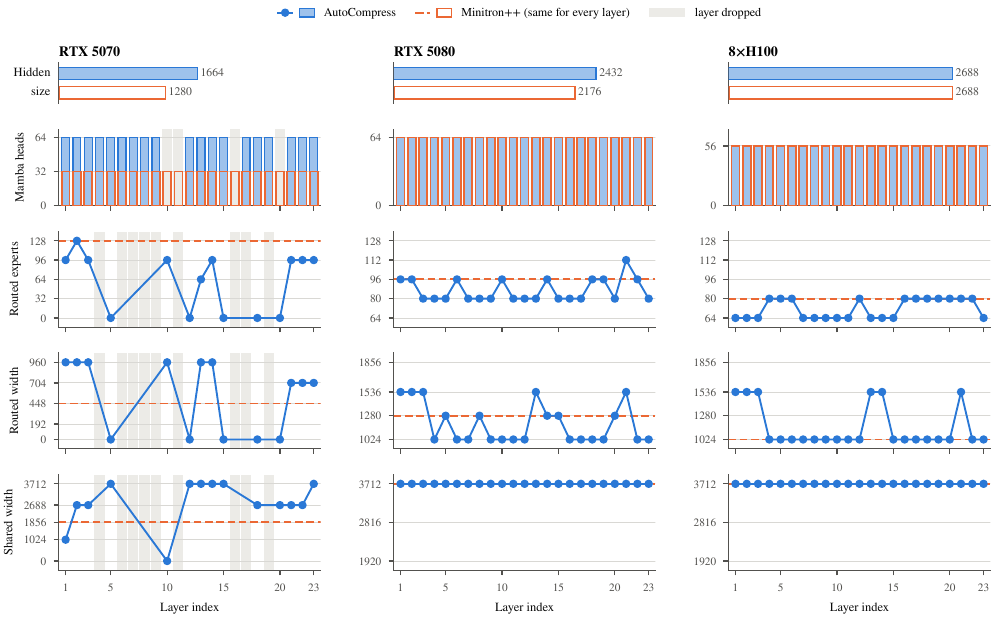}
\label{fig:layer-strips}
\end{figure*}

Figure~\ref{fig:layer-strips} shows every Mamba and FFN layer of the \method{} and Minitron students.

\section{Cost-Model Details}
\label{app:costmodel}

\subsection{Cost-model Accuracy}
\label{sec:validation}

\begin{figure*}[t!]
\centering
\includegraphics[width=\textwidth]{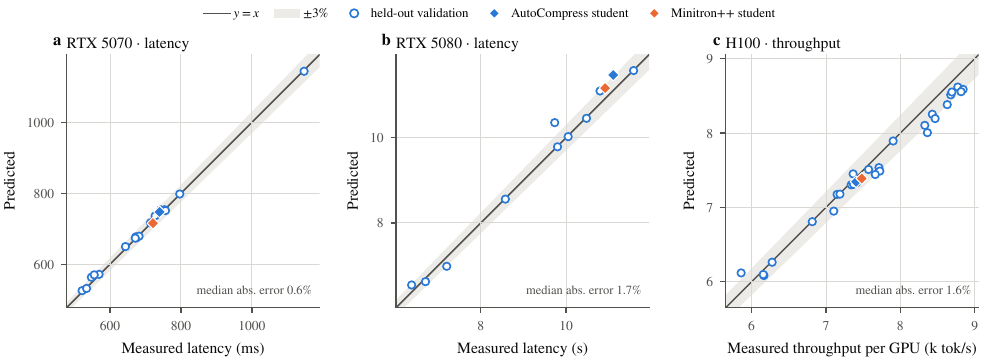}
\caption{\textbf{Cost-model accuracy}: predicted vs.\ measured end-to-end latency (RTX~5070, RTX~5080) and throughput per GPU (H100) on held-out architectures, including the searched ones.}
\label{fig:costmodel-accuracy}
\end{figure*}

We test the timing models on complete homogeneous and heterogeneous architectures with different widths and depths, none used to construct the layer timing tables. Figure~\ref{fig:costmodel-accuracy} compares predictions with serving measurements.

\begin{figure}[t]
\centering
\includegraphics[width=\columnwidth]{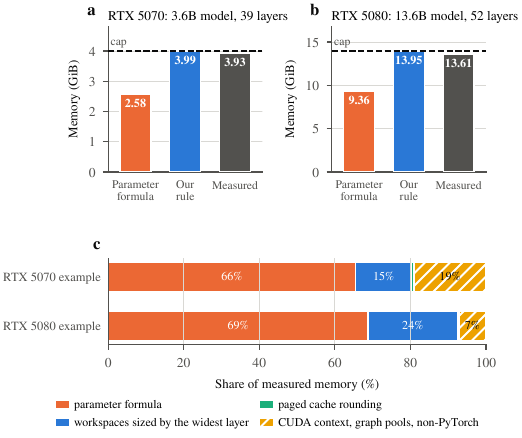}
\caption{\textbf{A conventional memory formula misses about a third of the memory.}
(a,~b)~Memory predicted by a conventional formula (weights at deployed precision, per-sequence KV cache and Mamba state, an MoE activation buffer) and by our memory cost model (\S\ref{sec:memory}), against measured vLLM memory: a 3.6B-parameter, 39-layer model on the RTX~5070 and a 13.6B-parameter, 52-layer model on the RTX~5080.
(c)~What the formula misses, as shares of measured memory.}
\label{fig:params-memory}
\end{figure}

For memory, we compare with an estimate based on weights at their deployed precision, request KV cache, and recurrent state, following the components commonly used in compression cost models~\citep{bercovich2024puzzle,bercovich2026nemotron}. Across six served RTX models, this estimate under-predicts memory by \prov{31--43}\%. Figure~\ref{fig:params-memory} includes the stated MoE activation buffer and shows one example on each GPU. The remaining allocations include prefill workspaces, other runtime buffers, CUDA-graph pools, page rounding, and non-PyTorch memory. Explicit accounting reduces the error by roughly an order of magnitude in these measurements.

\subsection{Profiling and training cost}
\label{sec:overhead}

Table~\ref{tab:measure-cost} reports the runs, timing-table entries, and GPU-hours needed to profile each deployment target. On RTX~5080, \prov{60} runs covering \prov{468} entries took approximately \prov{3.5} GPU-hours. H100 profiling used \prov{180} runs and \prov{24.8} GPU-hours, while RTX~5070 used \prov{96} runs and \prov{4.5} GPU-hours. Each short profiling model contains multiple configurations, allowing one run to contribute several layer timings.

\begin{table}[t!]
\centering
\footnotesize
\caption{Measurement cost of each contract's latency cost model (Table~\ref{tab:contracts}).}
\label{tab:measure-cost}
\begin{tabular}{@{}lccc@{}}
\toprule
\textbf{Contract} & \textbf{Runs} & \textbf{Table entries} & \textbf{GPU-h} \\
\midrule
Local coding (RTX~5080) & \prov{60} & \prov{468} & \prov{3.5} \\
Datacenter throughput (H100) & \prov{180} & \prov{408} & \prov{24.8}\\
On-device NPC (RTX~5070) & \prov{96} & \prov{660} & \prov{4.5} \\
\bottomrule
\end{tabular}
\end{table}


\section{Model Serving Details}
\label{sec:apx-serving}

Serving uses vLLM~0.27.1~\citep{kwon2023vllm}. Experts, shared experts, and the LM head use NVFP4; Mamba projections use FP8; attention uses BF16; and KV cache and Mamba state use FP8 and FP32, respectively. Timing and memory predictions use these deployed precisions. The H100 throughput target is evaluated at a fixed batch size, so a reduction in batch serving time also reduces completion time for those requests.

\section{Large-Scale Case Study}
\label{app:scale-details}

\begin{figure*}[!t]
\centering
\includegraphics[width=0.96\textwidth]{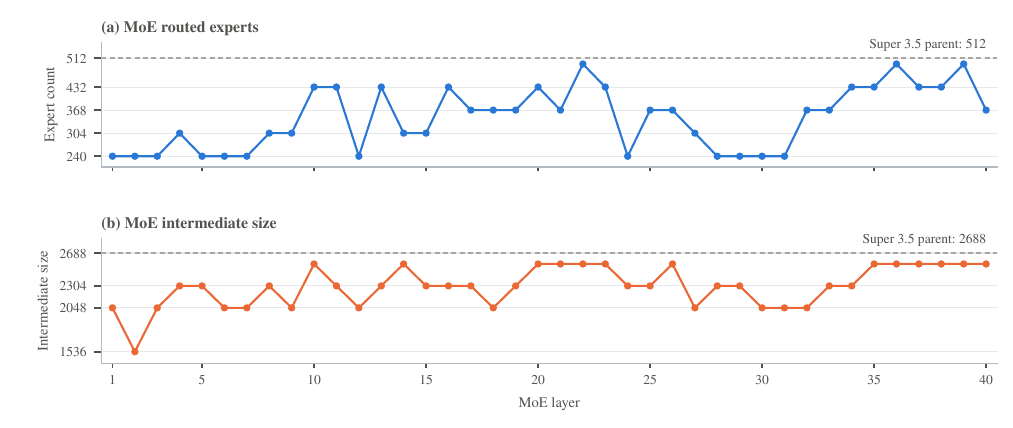}
\caption{\textbf{Final MoE architecture selected by \method{}.}
(a) MoE routed expert counts and (b) MoE intermediate sizes across all 40 MoE layers, ordered by backbone depth. Dashed lines show the parent's uniform dimensions: 512 routed experts and intermediate size 2688.}
\label{fig:scale-final-architecture}
\end{figure*}

Figure~\ref{fig:scale-final-architecture} shows the final architecture selected for the Nemotron 3.5 Super case study in \S\ref{sec:scale-case-study}. MoE routed expert counts range from 240 to 496, and MoE intermediate sizes range from 1536 to 2560, compared with 512 and 2688 in the parent. The model retains all 88 backbone layers, hidden size 4096, and shared-expert intermediate size 5376.

\begin{figure*}[!t]
\centering
\includegraphics[width=\textwidth]{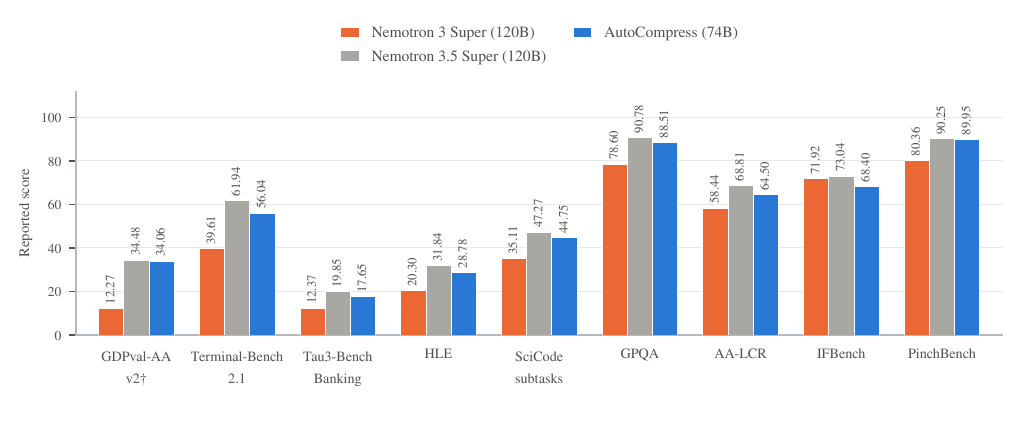}
\caption{\textbf{Benchmark scores after compression and distillation.}
Nemotron 3.5 Super and \method{}, with Nemotron 3 Super as a reference. The unweighted means across these nine metrics are 57.58, 54.74, and 45.44, respectively.
$\dagger$~GDPval-AA v2 is an index; the other scores are percentages.}
\label{fig:scale-quality}
\end{figure*}

Figure~\ref{fig:scale-quality} compares individual benchmark scores for \method{}, its Nemotron 3.5 Super parent, and Nemotron 3 Super. The unweighted mean covers nine reported metrics; GDPval-AA v2 is an index, while the remaining eight scores are percentages.

\end{document}